\RequirePackage{fix-cm}
\documentclass[twocolumn]{svjour3}          
\smartqed  
\usepackage{graphicx}
\usepackage{tabularx}
\usepackage{booktabs}
\usepackage{doi}
\usepackage{enumitem}
\usepackage{multirow} 
\usepackage{ifthen}
\usepackage{wrapfig}
\usepackage{amsmath}
\usepackage[ruled,vlined,linesnumbered]{algorithm2e}
\usepackage{makecell}
\usepackage[table]{xcolor}

\journalname{CGI2025} 
\begin{document}

\title{MAS-KCL: Knowledge Component Graph Structure Learning with Large Language Model-based Agentic Workflow}
\author{Yuan-Hao Jiang$^{\dagger}$ \and Kezong Tang$^{\dagger, *}$ \and Zi-Wei Chen$^{\dagger, *}$ \and Yuang Wei \and Tian-Yi Liu \and Jiayi Wu}
\institute{This paper has been accepted by \textit{CGI 2025: 42nd Computer Graphics International Conference} and recommended for publication in the SCI-indexed journal \textit{The Visual Computer}. The final published version is available at \texttt{https://doi.org/10.1007/s00371-025-03946-1} \\ \\$^{\dagger}$These authors contributed equally to this work. \\ \\ $^{*}$Corresponding authors: {tangkezong@jci.edu.cn} (K. Tang), \\ {ziwei.cs@foxmail.com} (Z.-W. Chen) \\ \\ Y.-H. Jiang \at Shanghai Institute of Artificial Intelligence for Education, East China Normal University, Shanghai, China \\ School of Computer Science and Technology, East China Normal University, Shanghai, China \\ Lab of Artificial Intelligence for Education, East China Normal University, Shanghai, China \\ Graduate School, Shanghai Jiao Tong University, Shanghai, China \and K. Tang \and Z.-W. Chen \at School of Information Engineering, Jingdezhen Ceramic University, Jingdezhen, China \and Y. Wei \at School of Computer Science and Technology, East China Normal University, Shanghai, China \\ Lab of Artificial Intelligence for Education, East China Normal University, Shanghai, China \\ Shanghai Institute of Artificial Intelligence for Education, East China Normal University, Shanghai, China \\ School of Computing, National University of Singapore, Singapore \and T.-Y. Liu \at School of Computer, Jiangsu University of Science and Technology, Zhenjiang, China \and J. Wu \at Department of Educational Information Technology, East China Normal University, Shanghai, China \\ Lab of Artificial Intelligence for Education, East China Normal University, Shanghai, China}
\date{}

\maketitle

\begin{abstract}
Knowledge components (KCs) are the fundamental units of knowledge in the field of education. A KC graph illustrates the relationships and dependencies between KCs. An accurate KC graph can assist educators in identifying the root causes of learners' poor performance on specific KCs, thereby enabling targeted instructional interventions. To achieve this, we have developed a KC graph structure learning algorithm, named MAS-KCL, which employs a multi-agent system driven by large language models for adaptive modification and optimization of the KC graph. Additionally, a bidirectional feedback mechanism is integrated into the algorithm, where AI agents leverage this mechanism to assess the value of edges within the KC graph and adjust the distribution of generation probabilities for different edges, thereby accelerating the efficiency of structure learning. We applied the proposed algorithm to 5 synthetic datasets and 4 real-world educational datasets, and experimental results validate its effectiveness in learning path recognition. By accurately identifying learners' learning paths, teachers are able to design more comprehensive learning plans, enabling learners to achieve their educational goals more effectively, thus promoting the sustainable development of education.
\keywords{Graph structure learning \and Knowledge component \and Large language model \and Multi-agent system \and AI agent}
\end{abstract}

\section{Introduction}

\begin{figure*}[t]
  \centering
  \includegraphics[width=\linewidth]{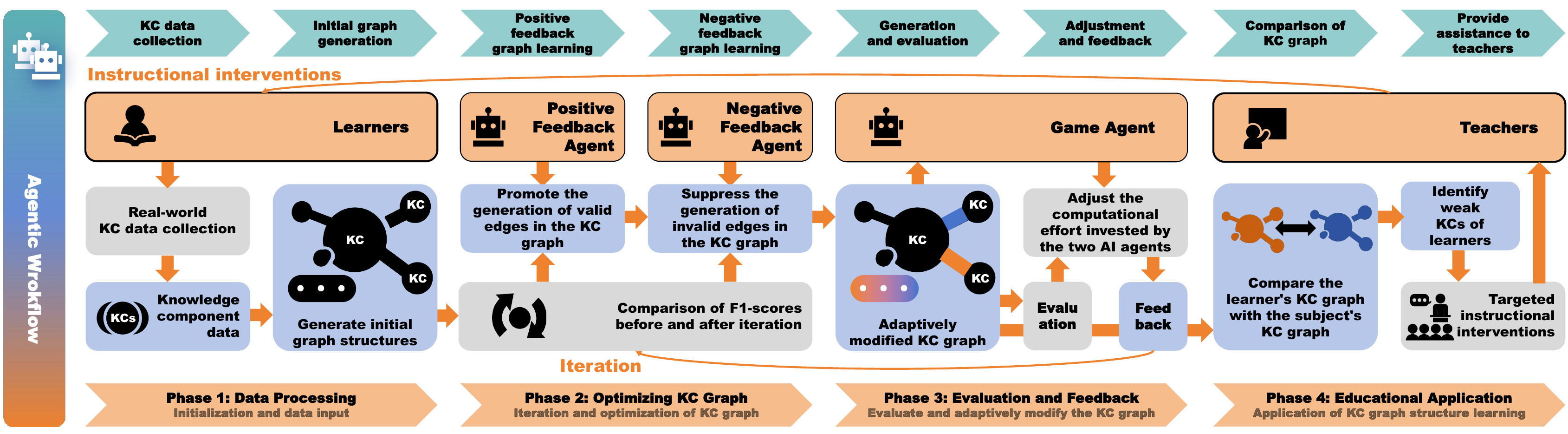}
  \caption{The proposed agentic workflow for KC graph structure learning. This framework injects the emergent intelligence of large language models into the designed multi-agent system, enabling it to extract KCs from real-world learning data and construct KC graphs. The agentic workflow facilitates deep collaboration among learners, teachers, and AI agents, assisting teachers in identifying the root causes of learners' poor performance at the KC level, thereby providing support for the development of targeted instructional interventions.}
  \label{fig:abstract}
\end{figure*}

\sloppy

The increasing global emphasis on sustainable development has brought quality education to the forefront, as highlighted by the United Nations Educational, Scientific and Cultural Organization (UNESCO) and its 17 Sustainable Development Goals (SDGs), which include goals, such as equitable education, sustainable communities, and reduced inequalities. Among the key pillars of educational development are inclusivity, resource accessibility, educational technology, and sustainability. Educational technology, in particular, leverages digital tools and information systems to enhance interactivity and broaden access to learning. Its integration has transformed teaching and learning processes \cite{yildiz_current_2020,berrett_administrator_2012}. However, traditional digital education often lacks sufficient personalization and coherent instructional design, leading to cognitive overload. Addressing these limitations requires a deeper understanding of the structural relationships among knowledge components (KCs)—commonly referred to as learning paths \cite{liu_educational_2024}.

Learning paths capture the dependencies among knowledge components and play a crucial role in guiding learners' progression. Consequently, the completeness and hierarchical organization of knowledge component structures are widely recognized as key indicators of learner’ proficiency \cite{wadouh_effect_2014,spector_methodology_2006}. Recent studies have employed clustering methods to classify learners based on their knowledge structures, uncovering links between structure patterns and learning outcomes \cite{he_investigating_2023}. Graph structures \cite{meng_index-based_2022,meng_counting_2024}, particularly directed acyclic graphs \cite{meng_survey_2024}, offer an effective means to encode such educational data. In these graphs, nodes denote individual knowledge components and edges define directed dependencies, thereby visualizing the logical flow of learning.

To address the inherent complexity of real-world knowledge structures, we propose a multi-agent system for graph structure learning specifically tailored for learning path recognition. This system leverages a Differential Evolution (DE) algorithm augmented with dynamic population control, which is guided by Artificial Intelligence (AI) agents through a bidirectional feedback mechanism. Experimental evaluations demonstrate that our approach significantly improves the accuracy of learning path identification.  The main contributions are as follows:

\begin{itemize}[label=\textbullet]
 \item {We propose a Multi-Agent System for Knowledge Component Graph Structure Learning (MAS-KCL), as illustrated in Figure~\ref{fig:abstract}. The effectiveness of the proposed method is validated using real-world educational datasets.}

 \item {We develop a large language model (LLM)-powered agentic workflow for MAS-KCL, where the integration of LLMs effectively facilitates the structural learning of KC graphs.}

\item {The KC graphs generated by MAS-KCL capture causal relationships among KCs within a specific subject, offering data-driven support for instructional interventions. For example, if a learner struggles with multiplication, the root cause may lie in an insufficient understanding of addition-related KCs. The proposed method provides valuable insights for educational practice.}
\end{itemize}

\section{Related Work}

\subsection{Causal Graph}

When recognizing learning paths, it is essential to determine the relationships among KCs within a KC graph \cite{liu_educational_2024}. Most KCs exhibit causal dependencies, which can be effectively represented using a causal graph—a Directed Acyclic Graph (DAG) that models the connections between phenomena and their underlying causes. A method for constructing causal knowledge networks based on Bayesian Networks (BNs) has been proposed, leveraging causal relationships to build these structures \cite{wei_enhancing_2024}. 

The application of causal graphs in educational research has been widely acknowledged, offering key advantages for causal inference and facilitating empirical investigation \cite{feng_introduction_2024}. Their value lies in enabling explicit causal reasoning, supported by tools that assist researchers in uncovering causal links, as demonstrated in educational technology studies \cite{weidlich_causal_2024}. {It is important to clarify that the concept of a causal graph is closely related to the KC graph. A causal graph is defined as a graph composed of a set of nodes and edges, where the edges indicate causal relationships between elements in the node set. The KC graph further constrains the node set to represent knowledge components specifically.}

\subsection{Knowledge Component Graph Structure Learning}
{KC graph structure learning has become a central topic in Intelligent Tutoring Systems (ITS). Early studies relied on expert-defined prerequisite structures, typically modeled using BNs, where domain knowledge is encoded as probabilistic dependencies \cite{conati2002using,carmona2005introducing}. To enhance flexibility and reduce reliance on expert priors, later work proposed data-driven approaches that learn conditional probabilities directly from learners’ data \cite{desmarais2006learned}. These methods infer KC dependencies by analyzing the temporal evolution of knowledge mastery, comparing patterns predicted by learner models \cite{desmarais2006learned,piech2015deep}. Dynamic Bayesian Networks (DBNs) extended this line by embedding causal relations between KCs \cite{kaser2017dynamic}, though they suffer from parameter explosion when modeling multiple prerequisites. To improve scalability and interpretability, the Embedding Prerequisite Relationships in Student Modeling (E-PRISM) framework introduces interpretable parameters and infers causal dependencies from learning trajectories \cite{allegre2023discovering}, while more recent research applies interventional and counterfactual reasoning to extract explainable KC relationships \cite{wei_enhancing_2024}.}

{Moreover, data-driven approaches have also adopted Reinforcement Learning (RL)  \cite{sun2024crowd,li2021memorypath} and Graph Neural Networks (GNNs) \cite{senior2025graph,yin_improving_2022} for KC graph learning. GNNs are effective at capturing structural patterns but lack interpretability and causal representation, and are often limited to static graphs. RL-based methods treat graph construction as a sequential decision process, aligning with the dynamics of learning, but require careful reward design and struggle with sparse feedback. To address these limitations, we propose MAS-KCL, a multi-agent system driven by LLMs. MAS-KCL employs an agentic workflow where LLM-based agents iteratively generate and refine KC graphs through interpretable interactions. A bidirectional feedback mechanism enables agents to assess edge importance and dynamically adjust generation probabilities, leading to efficient, interpretable, and causally grounded KC structure learning.}

\subsection{AI Agent for Education}

{The Transformer architecture \cite{chen2025dsts,wang2025temporal,huang2023transmrsr}, driven by the attention mechanism \cite{zhou2025gamnet}, has underpinned the rapid evolution of LLMs \cite{sheng2025synthetic}, which are increasingly applied in educational contexts. LLMs can generate personalized learning plans tailored to learners' habits and backgrounds \cite{wang_adaptive_2021,li_interpretable_2025}, and support external tools through iterative dialogue \cite{zhuang_toree_2024}. Frameworks such as LangChain have demonstrated the capability to automatically generate contextualized mathematical multiple-choice questions with improved accuracy \cite{li_large_2024}.} Recent studies have explored how LLMs and adaptive learning models serve as interfaces for human-machine interaction in personalized education \cite{wen_ai_2024}, and how Multimodal Large Language Models (MLLMs) can provide synchronized visual content to enhance mathematics learning experiences \cite{jiang_synchronizing_2024}.

Parallel to this development, LLMs-based AI agents have emerged as versatile tools in educational systems. By integrating web services and modeling learner characteristics, agents can infer learning domains and facilitate intelligent information retrieval \cite{li_artificial_2021}. Multi-agent systems (MASs), including those based on von Neumann architectures, have shown effectiveness in enhancing learners’ learning and supporting teaching processes \cite{jiang_ai_2024}. While MASs cannot replace human roles entirely, they promote multi-agent collaboration and learner interaction, improving learning outcomes \cite{shi_mitigating_2025}. In this work, we adopt a MAS framework to optimize the learning of KC graph structures.

\section{MAS-KCL for KC Graph Structure Learning}

{\subsection{LLM-based Multi-Agent System}}

\begin{figure*}[t]
  \centering
  \includegraphics[width=\linewidth]{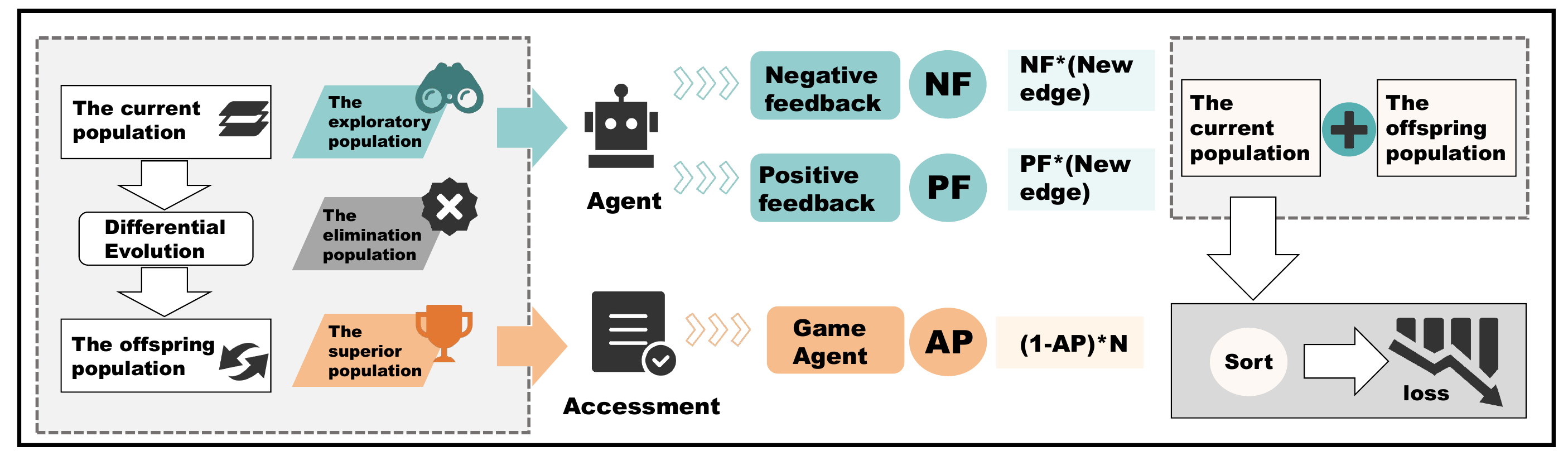}
  \caption{Game agent and feedback agents. The diagram displays the interaction between the Game Agent and the population, as well as the probability of adding new edges in the bidirectional feedback mechanism generated by the Feedback Agents.}
  \label{fig:2}
\end{figure*}

{This section introduces the multi-agent system composed of AI agents, which dynamically control parameters to deliver decision outputs during the algorithm’s iterative process. These agents interact with MAS-KCL by influencing the optimization process through decisions continuously updated via LLM calls. The overall interaction between agents and the MAS-KCL-controlled population is illustrated in Figure~\ref{fig:2}. In addition, the pseudocode of the proposed method is provided in Algorithm~\ref{alg:mas-kcl}.}

\begin{algorithm}[t]
\caption{Multi-Agent System for Knowledge Component Graph Structure Learning} \label{alg:mas-kcl}
\KwIn{$AP$ (Ambient Pressure), $OD$ (Optimization Direction)}
\KwOut{best individual and its $loss$}

Initialize the population randomly\;

$EF \leftarrow 0$\;

\While{$EF <$ max$EF$}{

    Generate offspring population by DE operator\;

    $OffPop \leftarrow sort([OffPop, OD])$\;
    \tcp{Sort individuals in the offspring population according to $OD$.}

    $OffPop \leftarrow \{( AP)*N[P_{SUP}], (1-AP)*N[P_{EX}],$ $N[P_{EL}] \}$\;
    \tcp{Divide the offspring population into superior and exploratory, and elimination sub-populations.}

    Establish bidirectional feedback mechanism\;

    $P_{EX} \leftarrow sort([P_{PFO},P_{NFO}]$\;
    \tcp{Sort individuals within the exploratory sub-population.}

    \ForEach{$i \in \{P_{PFO}, P_{NFO}\}$}{

        $NewEdges \leftarrow OffPop(1,i).dec$ - $Pop(1,i).dec$\;

        $count\_ones \leftarrow count\_ones + NewEdges$\;
        \tcp{Record the new edges of the exploratory sub-population for the current population.}

    }

    Multiply the vector of the newly added edge by the Positive Factor\;

    Multiply the vector of the newly added edge by the Negative Factor\;

    \ForEach{$i \in \{P_{PFO}, P_{NFO}\}$}{

        $OffPop(1,i).dec$ is modified by adding or deleting edges based on $count\_ones$\;

    }

    $sort \{Fitness(Pop), Fitness(OffPop)\}$\;
    \tcp{The fitness of individuals in the offspring population was compared with those in the current one.}


    $Pop(replace) \leftarrow OffPop(replace)$\;
    \tcp{Sub-population with lowest fitness is eliminated, and the remaining sub-populations are retained.}

    $Pop([P_{SUP}],[P_{EX}] \leftarrow OffPop\{(AP)*N$, $(1-ap)*N\}$\;
    \tcp{A portion of uperior population will enter the next iteration. The exploratory population will undergo a bidirectional feedback mechanism operation.}

    $AP \leftarrow$ Game Agent decision\;

    $PF \leftarrow$ Positive Feedback Agent decision\;

    $NF \leftarrow$ Negative Feedback Agent decision\;

    $FE \leftarrow FE + N$\;

}

\Return best individual and its $loss$\;
\end{algorithm}

{Specifically, after each iteration of the algorithm, the Game Agent is invoked using prompt engineering. It is asked to determine whether to adjust the Ambient Pressure ($AP$) parameter in MAS-KCL—based on the magnitude of $loss$ change observed in the current iteration—and to what extent. The decision is returned in a structured JSON format, which includes both the decision outcome and its reasoning. Before the next iteration begins, the value of $AP$ is updated accordingly, allowing the algorithm to remain dynamically aligned with the intermediate decisions made by the Game Agent, as described in Algorithm~\ref{alg:mas-kcl}. The value of the $AP$ parameter determines the proportion of elite individuals retained for the next iteration. It reflects the dynamic allocation of computational effort between exploration and exploitation.}

{Similarly, the Positive Feedback Agent and Negative Feedback Agent are invoked at the end of each iteration. Based on the current $loss$ change and parameter status, they make decisions regarding the activation and adjustment of the positive and negative feedback mechanisms, respectively. These decisions take effect in the subsequent iteration and are passed to MAS-KCL through dynamic parameter control, as described in Lines 20–21 of Algorithm~\ref{alg:mas-kcl}. The Positive Feedback Agent focuses on identifying edges that contribute to increased $loss$ and aims to eliminate these low-quality edges broadly across the population. In contrast, the Negative Feedback Agent is responsible for identifying edges associated with decreased $loss$ and promotes the widespread addition of these valuable edges throughout the population. Together, these agents collaborate to enable efficient learning of the KC graph structure, thereby supporting teachers in making informed instructional interventions, as illustrated in Figure~\ref{fig:abstract}. Further details on the interaction between the multi-agent system and the MAS-KCL framework are provided in the following Section~\ref{sec:The_Structure_of_the_Proposed_MAS-KCL}.}

\subsection{The Structure of the Proposed MAS-KCL}\label{sec:The_Structure_of_the_Proposed_MAS-KCL}

To improve the balance between exploration and exploitation, the proposed MAS-KCL adopts a multi-sub-population strategy that partitions the population into superior, exploratory, and elimination sub-populations. This structure allows certain individuals to exploit known promising regions while others explore new areas, enhancing global search capacity and avoiding premature convergence. The algorithm evaluates each individual using a fitness score $F$, with the $loss$ defined as $1 - F$ \cite{liu_educational_2024}. A higher $F$ indicates better structural alignment and, consequently, a lower $loss$, reflecting a more effective solution.

The main process of the algorithm is detailed in Algorithm~\ref{alg:mas-kcl}. {The procedure begins with the initialization of a random population and the setting of the Function Evaluation ($FE$) counter to zero, as shown in Lines 1-2 of Algorithm~\ref{alg:mas-kcl}. In each iteration, DE is employed to generate an offspring population, which is then sorted according to a predefined optimization direction, as described in Lines 4-5. The offspring are ranked in ascending order of $loss$ and divided into three sub-populations, as outlined in Lines 6-7. The proportion of the superior sub-population is controlled by the ambient pressure parameter $AP$, where a higher $AP$ preserves more elite individuals, and a lower value encourages exploration. The exploratory sub-population, sized as $(1-AP)\cdot N$, is further refined through a bidirectional feedback mechanism, as implemented in Lines 7-15 of Algorithm~\ref{alg:mas-kcl}.}

\begin{figure}[t]
  \centering
  \includegraphics[width=\linewidth]{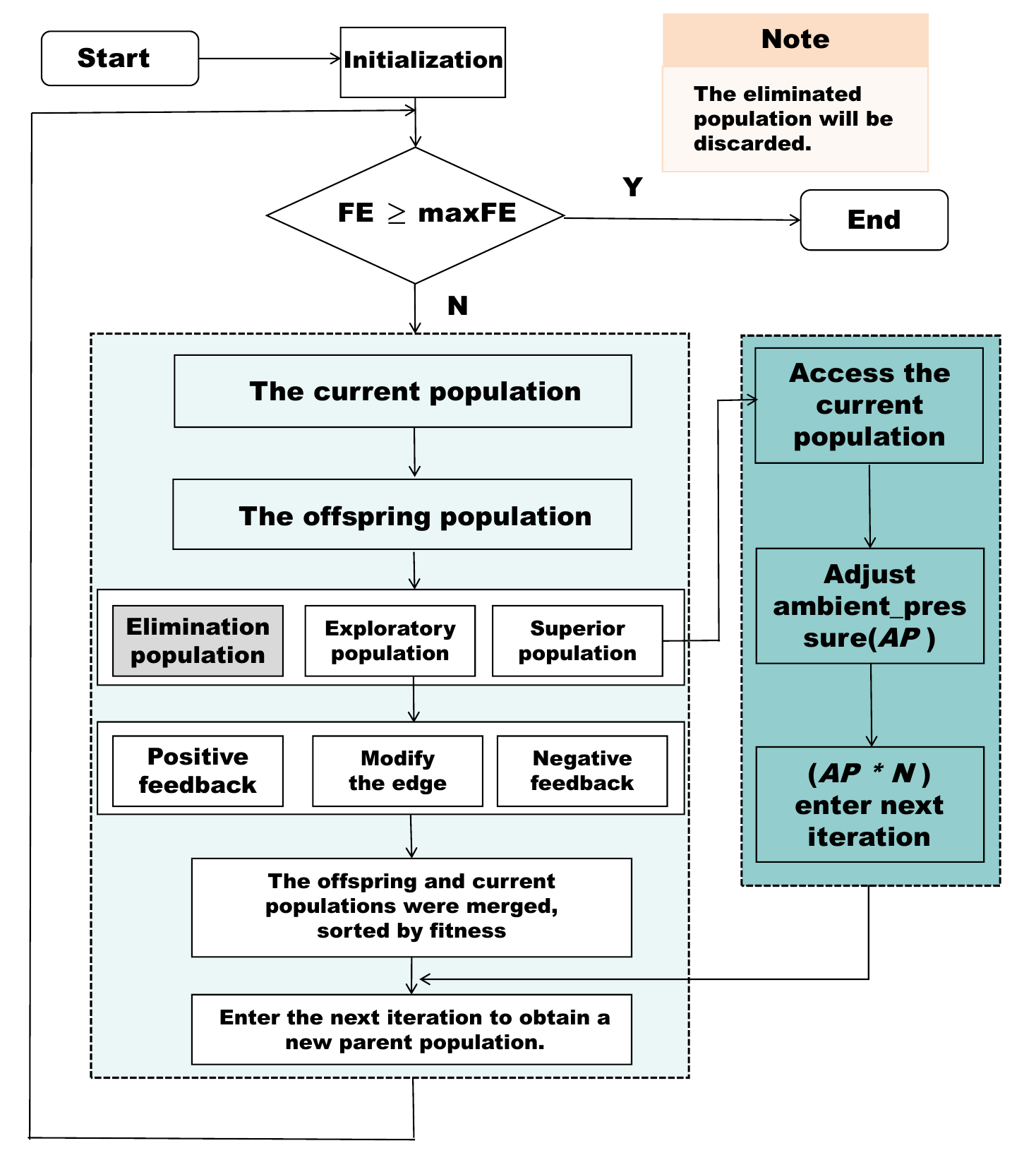}
  {\caption{KC graph structure learning search process. Figure 1 shows the search mechanism for the graph KC Structure based on multi-sub-population collaboration.}\label{fig:1}}
\end{figure}

{Within this feedback mechanism, individuals are updated based on their sub-population assignments: $P_{PFO}$ (high-fitness exploratory individuals) and $P_{NFO}$ (remaining individuals) influence the addition or deletion of edges through positive and negative factors ($PF$, $NF$). These structural changes are quantified using the metric $count\_ones$. The modified exploratory sub-population is then merged with the superior sub-population and passed to the next generation, as shown in Line 17 of Algorithm~\ref{alg:mas-kcl}. Following this, both parent and offspring populations are re-ranked, and the least fit individuals are eliminated to maintain selection pressure and solution quality, as described in Lines 16-17. Then,  the best individuals are retained based on $AP$, and the remaining ones are refined through the feedback mechanism for the subsequent iteration, as described in Lines 18-21 of Algorithm~\ref{alg:mas-kcl}. This iterative process continues until $FE$ reaches the maximum Function Evaluation ($\max FE$), at which point the algorithm terminates and outputs the optimal individual, as shown in Lines 22-23. The structure learning process for the KC graph is illustrated in Fig.~\ref{fig:1}, and the fitness definition aligns with prior work \cite{liu_educational_2024}. }

\section{Results}

\subsection{Experimental Setup}

\subsubsection{Educational Datasets} \label{EducationalDatasets}

{The experiments in this study are mainly conducted using datasets from the NeurIPS CausalML Challenge and the XuetangX platform, both of which are suitable for the task of knowledge component graph structure learning. A total of 9 datasets are used in this research, including 5 synthetic datasets and 4 real-world datasets. Detailed descriptions of the datasets are provided below according to their respective sources.}

\begin{itemize}[label=\textbullet]

  \item { \textbf{Learning Path Recognition-Generate Datasets (LPR-GDs) \cite{gong2022neurips}: } The LPR-GDs series datasets are provided by the \textit{NeurIPS CausalML Challenge: Causal Insights for Learning Paths in Education}. This series includes five generated datasets—LPR-GD1 to LPR-GD5—which contain data on the causal relationships among KCs in the domain of mathematics. Although there exists a gap between synthetic and real-world data, the LPR-GDs datasets are well-suited for the preliminary validation and rapid testing of KC graph structure learning methods. }

  \item { \textbf{Learning Path Recognition-Real World Datasets (LPR-RWDs) \cite{gong2022causalml}: } The LPR-RWDs series datasets provided by \textit{Microsoft Research Cambridge} and \textit{Eedi}. The LPR-RWDs include three real-world datasets of varying scales: LPR-RWD, LPR-RWD1, and LPR-RWD2. These datasets are constructed from real-world educational data provided by Eedi, an online education company registered in England and Wales. The LPR-GDs series datasets represent learning data collected from real learners engaged in mathematics education, making them well-suited for thoroughly evaluating the generalizability of KC graph structure learning methods. }

  \item { \textbf{MOOCCubeX-Math \cite{yu2021math}: } XuetangX is one of the largest MOOC platforms in China. The MOOCCubeX dataset is collected on the XuetangX platform, which provides partially annotated and algorithmically predicted prerequisite relationships between concepts \cite{yu2021mooccubex}. By treating the concepts defined in MOOCCubeX as KCs, we can obtain real-world data suitable for the task of KC graph structure learning. Specifically, we extracted MOOCCubeX-Math from the 331,202 mathematics-related JSON records provided by MOOCCubeX and performed data cleaning based on the methodology in \cite{shokouhinejad2024node}, in order to further evaluate the generalizability of the proposed approach. }

\end{itemize}

\subsubsection{Baseline Algorithms}

{In this study, we compare our proposed algorithm, MAS-KCL, against four representative baseline algorithms: Multi-Stage multi-objective Evolutionary Algorithm (MSEA) \cite{tian_multistage_2021}, Golden Eagle Optimizer (GEO) \cite{mohammadi_golden_2021}, Elite Evolution Strategy-based Harris Hawk Optimization algorithm (EESHHO) \cite{heidari_harris_2019}, and Adaptive Geometry Estimation-based Many-Objective Evolutionary Algorithm II (AGE-MOEA-II) \cite{panichella_improved_2022}. The selected methods exhibit diverse characteristics, making them suitable for a comprehensive comparison and evaluation of the proposed MAS-KCL from multiple perspectives. Our selection was guided by the recommendations in \cite{liu_educational_2024}, from which MSEA, GEO, and EESHHO were adopted. Additionally, we included AGE-MOEA-II, a recently proposed algorithm with strong theoretical foundations, to further validate the effectiveness of the proposed LLM-driven approach.}

It is worth noting that the parameter settings for all comparison algorithms were either derived from their original papers or based on the default configurations provided by the Platform for Evolutionary Multi-objective Optimization (PlatEMO) \cite{8065138}. The hyperparameters of the comparison algorithms are listed in Table \ref{tab:1}. It should be noted that AGE-MOEA-II \cite{panichella_improved_2022} is not included in Table \ref{tab:1} because it does not contain adjustable parameters.

\begin{table*}[t]  
  \centering 
  \begin{tabularx}{\textwidth}{>{\centering\arraybackslash}p{3cm}|>{\centering\arraybackslash}X} 
    \toprule
    \textbf{Algorithm} & \textbf{Parameters}  \\  
    \midrule
    GEO\cite{mohammadi_golden_2021} & $AP_{min}$ = 0.5, $AP_{max}$ = 2.0, $CP_{min}$ = 1.0, $CP_{max}$ = 0.5 \\  
    EESHHO\cite{heidari_harris_2019} & $U_b$ = 1, $L_b$ = 0 \\  
    MSEA\cite{tian_multistage_2021} & $F_{max} =1$, $F_{min}$ = 0   \\  
    MAS-KCL & $PF$ = 0.6, $NF$ = 0.4, $AP$ = 0.4   \\  
    \bottomrule
  \end{tabularx}
  \caption{Parameter settings of all algorithms. This table presents the algorithm parameters used in the simulation experiments.}  
  \label{tab:1}  
\end{table*}

\subsubsection{Implementation Details}

{The experiments were primarily conducted on a machine equipped with an Intel i5-13600KF @ 3.50GHz dual-core processor, 32GB of RAM, and an NVIDIA RTX 4070 GPU. The software environment included MATLAB 2022b and the PlatEMO platform \cite{8065138}. Additionally, the online tool \textit{Chiplot.online} was used for part of the data processing and visualization in this study. Following previous work, we employed the $loss$ metric to evaluate the performance \cite{liu_educational_2024}.}

\subsection{Comparison Experiments on Real-World Datasets} \label{sec:Comparison_Experiments_on_Real-World_Datasets}

{This subsection compares the MAS-KCL with four algorithms on real-world datasets. To reduce the impact of random factors, each algorithm was run at least three times for each dataset. The results are shown in Table \ref{tab:2}, where lower $loss$ indicates better performance. }

\begin{table*}[t]  
  \centering  
  \begin{tabularx}{\textwidth}
  {>{\centering\arraybackslash}m{2.1cm}
>{\centering\arraybackslash}m{2cm}
>{\centering\arraybackslash}m{0.01cm}
>{\centering\arraybackslash}m{1.8cm}
>{\centering\arraybackslash}m{1.8cm}
>{\centering\arraybackslash}m{1.8cm}
>{\centering\arraybackslash}m{2.5cm}
>{\centering\arraybackslash}m{2cm}
}
    \toprule
    \multicolumn{2}{c}{\textbf{Real-World Dataset}} & & \multicolumn{5}{c}{\textbf{Algorithm}}\\
    \cmidrule(r){1-2}
    \cmidrule(r){4-8}
    
    \textbf{Dataset} & \textbf{D} & & \textbf{MSEA} & \textbf{GEO} & \textbf{EESHHO} & \textbf{AGE-MOEA-II} & \textbf{MAS-KCL} \\  
    \midrule

    {MOOCubeX-Math} & {210} & &  {7.37e-1}
 {(5.35e-3)[-]} & {6.46e-1}
 {(4.08e-2)[-]} & {\underline{2.57e-1}}
 {(4.95e-2)[-]} & {7.28e-1}\ \ \ \ \ \  \ 
 {(6.15e-3) [-]} & {\textbf{2.51e-1}}
 {(7.61e-2)} \\  
 
    LPR-RWD & 6670&  & 4.84e-1
 (1.78e-3) [-] & 3.44e-1 
(3.95e-5)[-] & \underline{3.43e-1}
 (2.76e-4)[-] & 4.81e-1\ \ \ \ \ \  \ 
 (2.72e-3) [-] & \textbf{3.31e-1} 
(1.30e-3) \\  
    LPR-RWD1 & 1225 &  & 4.13e-1
 (4.36e-3)[-] & 2.72e-1
 (2.06e-3)[-] & \underline{2.71e-1}
 (1.59e-3)[-] & 4.11e-1 \ \ \ \ \ \  \ 
(4.16e-3) [-] & \textbf{2.23e-1}
 (4.23e-3) \\  
    LPR-RWD2 & 1225 & & 4.43e-1
 (8.90e-3)[-] & 3.34e-1
 (2.15e-4)[-] & \underline{3.21e-1}
 (1.12e-2)[-] & 4.42e-1\ \ \ \ \ \  \ 
 (1.04e-2) [-] & \textbf{2.65e-1}
 (1.40e-2) \\

    \midrule  
    +/=/- & - & & {0/0/4} & {0/0/4} & {0/0/4} & {0/0/4} & - \\ 
    
    \bottomrule
  \end{tabularx}
  {\caption{Comparison of $loss$ (Mean ± SD) for five algorithms on four real-world datasets. Symbols "+", "=", and "–" indicate performance relative to the proposed method; the best and second-best results are shown in bold and underlined, respectively.} \label{tab:2} }
\end{table*}

{Table \ref{tab:2} presents the results on four real-world datasets: MOOCCubeX-Math, LPR-RWD, LPR-RWD1, and LPR-RWD2, with problem sizes $D$ of 210, 6670, 1225, and 1225, respectively. The results for each algorithm are reported as the average $loss$ values obtained through multiple independent runs. As shown in this table, LPR-RWD exhibits the highest $loss$ among the three LPR-RWDs series datasets due to its largest problem size, making optimization more challenging. Meanwhile, the MOOCCubeX-Math dataset is relatively sparse, which leads to poorer performance across most algorithms compared to the other datasets. Overall, the EESHHO algorithm consistently delivers second-best performance across all datasets, while the proposed MAS-KCL achieves the best results in every case, demonstrating its effectiveness in the task of KC graph structure learning.}

\subsection{Generalization Comparison on Generated Datasets}

{To further evaluate the generalization capability of the proposed method, we conducted additional comparison experiments on the generated datasets. Specifically, MAS-KCL was compared against the best-performing baseline algorithm identified in Section~\ref{sec:Comparison_Experiments_on_Real-World_Datasets}. The results on five generated datasets—LPR-GD1 to LPR-GD5—are presented in Table~\ref{tab:3}. Consistent with the results observed on real-world datasets, MAS-KCL achieves the best performance across all generated datasets. On average, MAS-KCL reduces the $loss$ by 5.51\%, with the greatest improvement observed on LPR-GD3, where the reduction reaches 7.71\%. These results confirm the strong generalization ability of the MAS-KCL on both real-world and generated datasets.}

\begin{table*}[t]  
  \centering  
  
  \begin{tabularx}{\textwidth}{
>{\centering\arraybackslash}m{2.1cm}
>{\centering\arraybackslash}m{1.8cm}
>{\centering\arraybackslash}m{1.8cm}
>{\centering\arraybackslash}m{1.8cm}
>{\centering\arraybackslash}m{1.8cm}
>{\centering\arraybackslash}m{1.8cm}
>{\centering\arraybackslash}m{0.01cm}
>{\centering\arraybackslash}m{1.2cm}
>{\centering\arraybackslash}m{1.45cm}
}

    \toprule
    \multirow{2}{*}{\textbf{Algorithm}} & \multicolumn{5}{c}{\textbf{Dataset}} & & \multicolumn{2}{c}{\textbf{Statistic}} \\  
    
    \cmidrule(r){2-6}
    \cmidrule(r){8-9}
    &  \textbf{LPR-GD1} & \textbf{LPR-GD2} & \textbf{LPR-GD3} & \textbf{LPR-GD4} & \textbf{LPR-GD5} & & \textbf{Mean} & \textbf{SD}\\  
    \midrule
    MAS-KCL & 26.99 & 26.13 & 27.30 & 29.19 & 28.85 & & 27.69 & 1.29 \\
    {Baseline} & {32.81} & {31.00} & {35.01} & {33.87} & {33.30} & & {33.20} & {1.48} \\
    
    \midrule

{$\Delta loss$} & \textcolor{blue}{$\downarrow$}~{5.82} & \textcolor{blue}{$\downarrow$}~{4.87} & \textcolor{blue}{$\downarrow$}~{7.71} & \textcolor{blue}{$\downarrow$}~{4.68} & \textcolor{blue}{$\downarrow$}~{4.45} & & \textcolor{blue}{$\downarrow$}~{5.51} & \textcolor{blue}{$\downarrow$}~{0.19} \\

    \bottomrule
  \end{tabularx}
 
  { \caption{Comparison results on the generated datasets. We compare the baseline with the proposed MAS-KCL on five datasets (LPR-GD1 to LPR-GD5) in terms of $loss$ (\%). The symbol \textcolor{blue}{$\downarrow$} indicates the percentage reduction in $loss$ achieved by MAS-KCL compared to the corresponding baseline. For each method, the mean and standard deviation are reported. } \label{tab:3}}
\end{table*}

\subsection{Generalization Sensitivity of MAS-KCL to LLMs}

{Since MAS-KCL integrates LLMs to enhance its decision-making process, we conducted a comparative study to analyze its generalization performance under different LLM versions. As shown in Table~\ref{tab:4}, we compared two GPT models developed by OpenAI (GPT-4.0 and GPT-3.5) with two alternative LLMs (LLaMA-70B and Claude-3-7-Sonnet-20250219) on four real-world datasets. The results show that the GPT models consistently outperform the other LLMs across all datasets. Notably, GPT-3.5 performs better on smaller-scale datasets such as MOOCCubeX-Math, while GPT-4.0 exhibits superior performance as the problem size increases, as seen in the LPR-RWD dataset. On average, the use of GPT models reduces the $loss$ by 5.79\% compared to other LLMs, demonstrating their strong compatibility with the MAS-KCL framework.}

\begin{table*}[t] 
  \centering 
  
  \begin{tabularx}{\textwidth}{
>{\centering\arraybackslash}m{2cm}
>{\centering\arraybackslash}m{3.2cm}
>{\centering\arraybackslash}m{2cm}
>{\centering\arraybackslash}m{2cm}
>{\centering\arraybackslash}m{2.3cm}
>{\centering\arraybackslash}m{0.01cm}
>{\centering\arraybackslash}m{1.1cm}
>{\centering\arraybackslash}m{1.6cm}
}

    \toprule
    \multirow{2}{*}{\textbf{LLM}} & \multicolumn{4}{c}{\textbf{Dataset}} & & \multicolumn{2}{c}{\textbf{Statistic}} \\  
    
    \cmidrule(r){2-5}
    \cmidrule(r){7-8}
    &  {\textbf{MOOCubeX-Math}} & \textbf{LPR-RWD} & \textbf{LPR-RWD1} & \textbf{LPR-RWD2} & & \textbf{Mean} & \textbf{SD}\\  
    \midrule
    GPT-4.0 & {\underline{25.12}} & \textbf{33.08} & \textbf{22.31} & \underline{26.47} & & {\underline{26.75}} & {\underline{4.57}} \\
    GPT-3.5 & {\textbf{19.58}} & \underline{33.55} & \underline{23.14} & \textbf{26.23} & & {\textbf{25.63}} & {5.94} \\
    {Claude} & {48.49} & {33.91} & {25.44} & {30.16} & & {34.50} & {9.95} \\
    {LLaMA} & {27.99} & {33.93} & {25.88} & {30.04} & & {29.46} & {\textbf{3.43}} \\

    \midrule

{$\Delta loss$} & \textcolor{blue}{$\downarrow$}~{15.89} & \textcolor{blue}{$\downarrow$}~{0.61} & \textcolor{blue}{$\downarrow$}~{2.94} & \textcolor{blue}{$\downarrow$}~{3.75} &  & \textcolor{blue}{$\downarrow$}~{5.79} &\textcolor{blue}{$\downarrow$}~{1.21} \\

    \bottomrule
  \end{tabularx}

  { \caption{The $loss$ results of MAS-KCL integrated with different versions of LLMs on real-world datasets. For each dataset, the best-performing and second-best methods are highlighted in bold and with an \underline{underline}, respectively. The table also reports the mean and standard deviation for each method, with consistent formatting applied. The symbol \textcolor{blue}{$\downarrow$} indicates the average $loss$ reduction achieved by GPT models (GPT-4.0 and GPT-3.5) compared to other variants.} \label{tab:4}}
\end{table*}

{Based on these observations, we suggest the following: GPT-3.5 should be prioritized for small-scale datasets; GPT-4.0 is more suitable for large-scale scenarios to achieve optimal results. When GPT models are not available, LLaMA serves as a viable alternative, offering the lowest standard deviation among all models, which indicates better stability—albeit at the cost of some performance. These findings offer valuable insights into selecting LLMs for integration with MAS-KCL in different application contexts.}

\subsection{Ablation Experiments}

\begin{figure}[t]
  \centering
  \includegraphics[width=\linewidth]{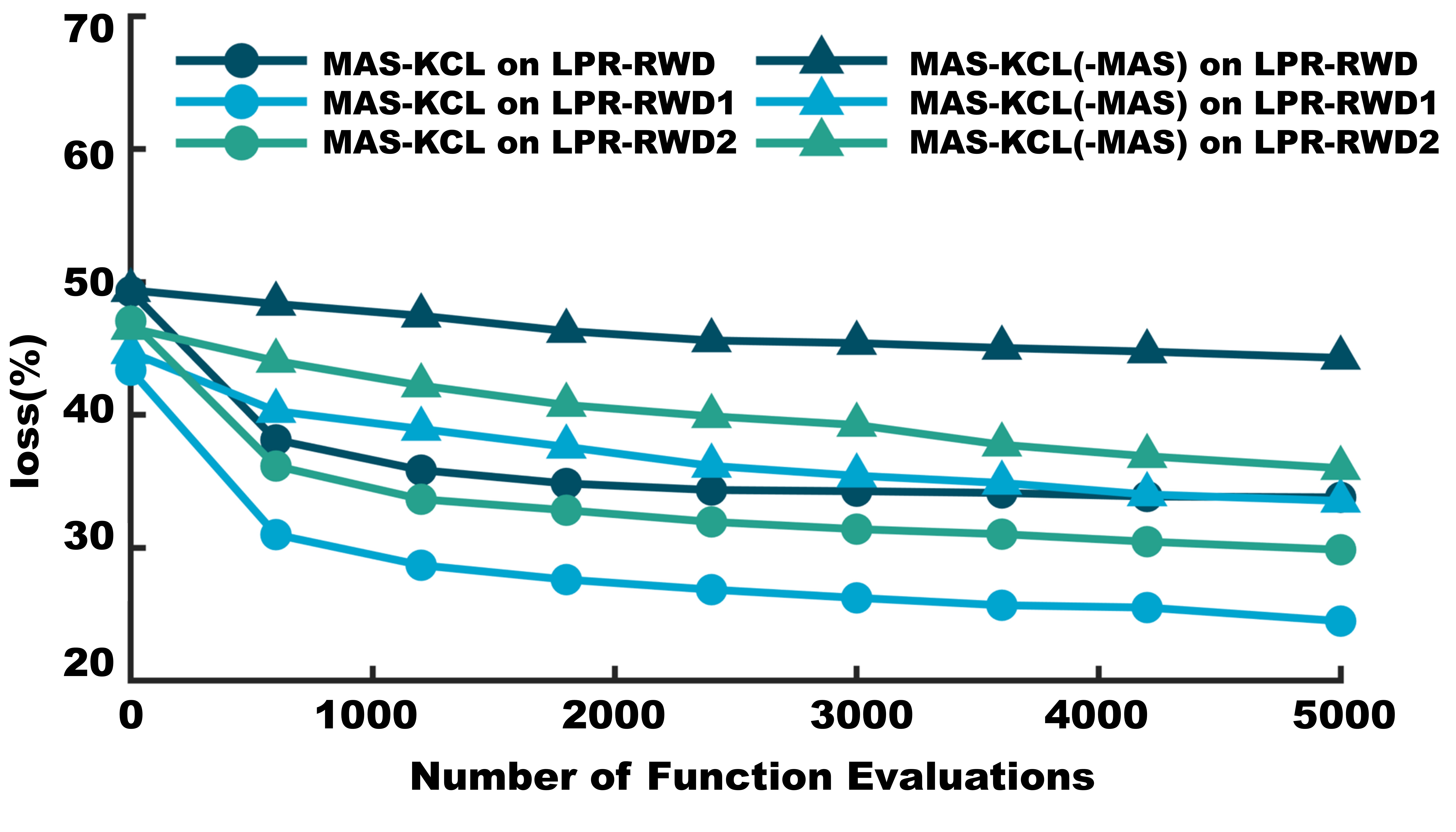}
  \caption{Results of ablation experiments on the multi-agent system component in MAS-KCL. These results illustrate the variation in $loss$ across different real-world educational datasets. Specifically, the comparison algorithm without the multi-agent system component is referred to as \textbf{MAS-KCL (-MAS)}, represented by triangles, while the standard \textbf{MAS-KCL} is depicted by circles.}
  \label{fig:4}
\end{figure}

To assess the contribution of the LLM-based multi-agent system to algorithm performance, we conducted an ablation study by removing the MAS component and evaluating the results on the real-world LPR-RWDs series datasets, as shown in Figure~\ref{fig:4}. Through iterative optimization, the standard MAS-KCL consistently achieved significantly lower $loss$ across all datasets compared to MAS-KCL(-MAS), which excludes the multi-agent system component. These findings demonstrate that integrating the multi-agent system component improves the efficiency of KC graph structure learning.

{To further investigate the impact of each agent, we conducted a more fine-grained ablation study, and the results are presented in Table~\ref{tab:ae}. We observed that the removal of any single agent led to an increase in $loss$, indicating that each agent contributes positively to the KC graph structure learning task. Specifically, removing the entire multi-agent system resulted in a 10.63\% increase in $loss$, while removing any single agent caused a rise of approximately 2\%–3\%. Among all agents, the removal of the Negative Feedback Agent (NFA) led to the largest increase in $loss$, suggesting that it plays the most critical role in algorithm performance. This is likely due to its ability to identify and eliminate low-quality edges between KCs. The Positive Feedback Agent (PFA) showed the second most significant impact, as it promotes the generation of high-quality edges through positive reinforcement. These findings provide strong evidence for the effectiveness of both the MAS as a whole and each of its individual components.}

\begin{table*}[t]  
  \centering  
  \begin{tabularx}{\textwidth}
  {>{\centering\arraybackslash}m{2.1cm}
>{\centering\arraybackslash}m{2cm}
>{\centering\arraybackslash}m{0.01cm}
>{\centering\arraybackslash}m{2cm}
>{\centering\arraybackslash}m{2cm}
>{\centering\arraybackslash}m{2cm}
>{\centering\arraybackslash}m{2cm}
>{\centering\arraybackslash}m{2.05cm}
}
    \toprule
    \multicolumn{2}{c}{\textbf{Real-World Dataset}} & & \multicolumn{5}{c}{\textbf{Algorithm}}\\
    \cmidrule(r){1-2}
    \cmidrule(r){4-8}
    
    \textbf{Dataset} & \textbf{D} & & {\textbf{MAS-KCL\ \ \ \ (-GA)}} & {\textbf{MAS-KCL\ \ \ \ (-PFA)}} & {\textbf{MAS-KCL\ \ \ \ (-NFA)}} & \textbf{MAS-KCL\ \ \ \ (-MAS)} & \textbf{MAS-KCL} \\  
    \midrule

    LPR-RWD & 6670&  & {\underline{3.39e-1}}& {3.42e-1} & {3.40e-1}& 4.43e-1& \textbf{3.31e-1} \\ 

    LPR-RWD1 & 1225 &  & {2.49e-1}& {\underline{2.46e-1}}& {2.55e-1}& 3.35e-1 & \textbf{2.23e-1}\\  
 
    LPR-RWD2 & 1225 & & {\underline{3.00e-1}}& {3.15e-1}& {\underline{3.15e-1}}& 3.60e-1& \textbf{2.65e-1}\\

    \midrule 
    Mean & - & & {2.96e-1} & {3.01e-1} & {3.03e-1} & {3.79e-1} & {2.73e-1} \\  
    {$\Delta loss$} & - & & \textcolor{blue}{$\uparrow$}~{2.30\%} & \textcolor{blue}{$\uparrow$}~{2.80\%} & \textcolor{blue}{$\uparrow$}~{3.03\%} & \textcolor{blue}{$\uparrow$}~{10.63\%} & - \\  
    
    \bottomrule
  \end{tabularx}
  {\caption{Ablation results of MAS-KCL with respect to the entire multi-agent system and each individual agent. Here, (-GA), (-PFA), (-NFA), and (-MAS) denote the removal of the Game Agent, Positive Feedback Agent, Negative Feedback Agent, and the entire LLM-based multi-agent system, respectively. For each dataset, the best-performing and second-best methods are highlighted in bold and with an \underline{underline}, respectively. \textcolor{blue}{$\uparrow$} indicates an increase in $loss$ compared to the standard MAS-KCL.} \label{tab:ae} }
\end{table*}

\subsection{Convergence Experiment of MAS-KCL}

\begin{figure*}[t]
  \centering
  \includegraphics[width=\linewidth]{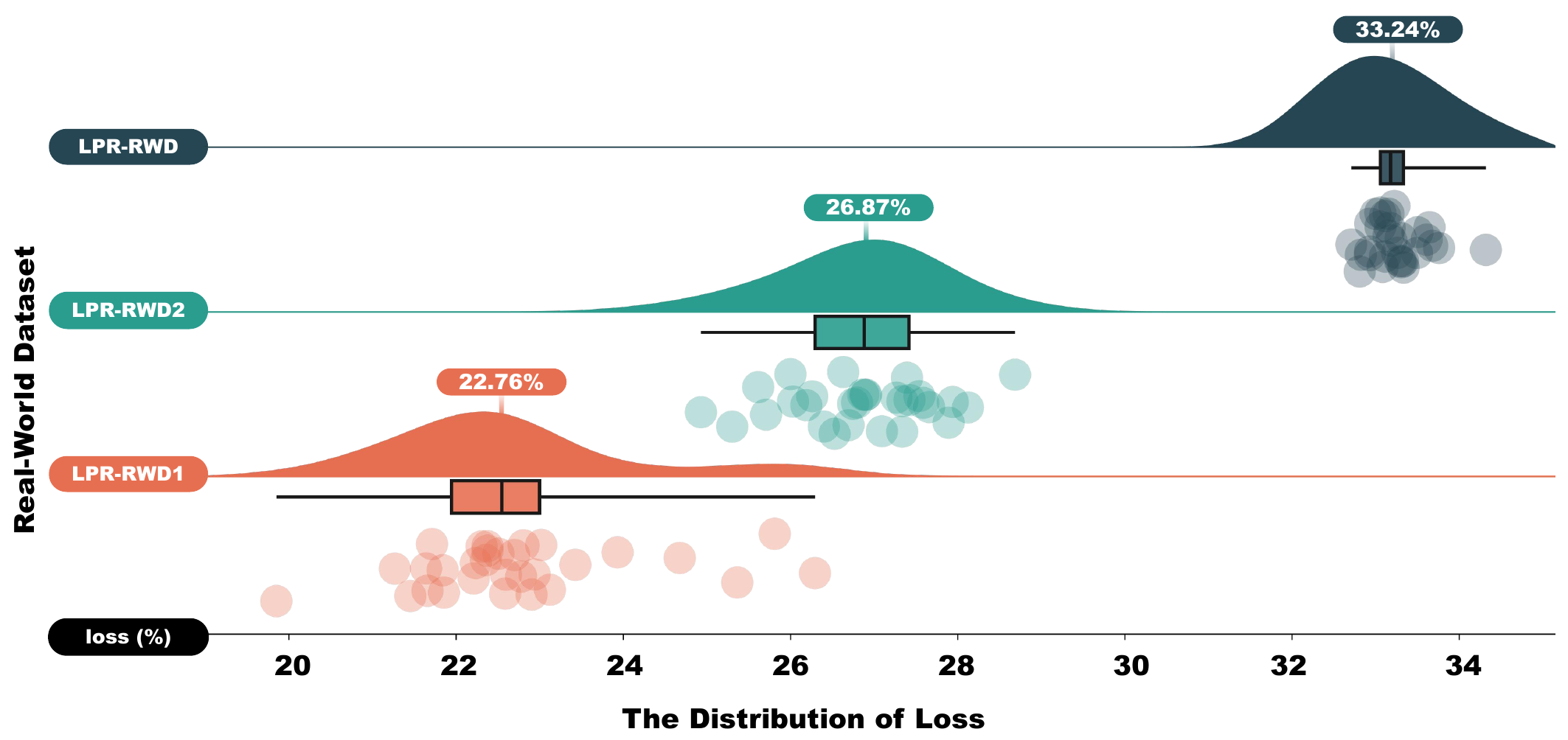}
  {\caption{Convergence analysis,which shows the convergence behavior and loss distribution of MAS-KCL. The centralized distribution of these results validates the strong convergence capability of the proposed method.} \label{fig:5}}
\end{figure*}

This subsection analyzes the convergence of the proposed MAS-KCL algorithm, which is a key measure of an algorithm's stability and effectiveness. In practical applications, convergence ensures that, after sufficient computation or iteration, the algorithm yields stable and consistent results. The datasets used in the experiments include LPR-RWD, LPR-RWD1, and LPR-RWD2. The results are presented in Figure \ref{fig:5}, which shows the convergence behavior and $loss$ distribution of MAS-KCL. These results were derived from 30 independent runs for each dataset. The boxplots below the x-axis illustrate the distribution of the 30 runs, while the median $loss$ for each problem is indicated as the central value. The ranking of $loss$ from highest to lowest is consistent with earlier results: LPR-RWD, LPR-RWD2, and LPR-RWD1.

For the LPR-RWD, the $loss$ distribution is dense, despite the dataset's high dimensionality contributing to its relatively high $loss$. The clear distribution patterns in Figure~\ref{fig:5} demonstrate the stability and convergence of MAS-KCL across different datasets.

\section{Conclusion}
To improve the efficiency and accuracy of KC graph structure learning and support learners in achieving their learning goals, we propose the Multi-Agent System for Knowledge Component graph structure Learning (MAS-KCL). MAS-KCL incorporates multi-agent collaboration and a bidirectional feedback system to optimize KC graph during iterations by dividing the population based on $loss$ values. Experiments demonstrated the proposed algorithm's effectiveness on real datasets. By enabling rapid and accurate KC graph structure learning, the MAS-KCL algorithm optimizes teaching strategies, enhances learning outcomes, improves the quality of educational development, and ensures sustainable progress in education.

  \vspace{3em}

\footnotesize
  \noindent \textbf{Conflicts of Interest} The authors declare no conflict of interest.

\vspace{0.5em}
\noindent \textbf{Acknowledgments} This work was supported in part by the Special Foundation for Interdisciplinary Talent Training in "AI Empowered Psychology / Education" of the School of Computer Science and Technology, East China Normal University (2024JCRC-03), and the Doctoral Research and Innovation Foundation of the School of Computer Science and Technology, East China Normal University (2023KYCX-03), and the Outstanding Doctoral Research Program for Advancing Academic Innovation at East China Normal University, with the title of "Research on the Generation and Mechanism of Visual Cues in Multimodal Dialog-Based Mathematics Tutoring", and the National Innovative Training Program for College Students of Jingdezhen Ceramic University (202410408028), and the Jiangxi Provincial Innovative Training Program for College Students of Jingdezhen Ceramic University (S202410408033). This work was carried out while Yuan-Hao Jiang was participating in an inter-university joint student program at Shanghai Jiao Tong University. The authors would like to express their special thanks to Xiaobao Shao and Hanglei Hu from East China Normal University for their valuable assistance. It is important to note that the proposed MAS-KCL is an algorithm enhanced with large language models (LLMs) to improve decision-making capabilities. In this study, four different LLMs were used in the experiments: OpenAI GPT-4.0, OpenAI GPT-3.5, LLaMA-70B, and  Claude-3-7-sonnet-20250219. No images in this study were generated using LLMs.

  \vspace{0.5em}
  \noindent \textbf{Data Availability} All datasets used in this paper are publicly available and can be accessed through open channels upon request. Detailed descriptions of all datasets and their access links can be found in the Educational Datasets Section of this paper.
  
\normalsize

\bibliographystyle{spmpsci}
\bibliography{sample-base} 


\end{document}